\documentclass[letterpaper, 10 pt, conference]{ieeeconf}  

\IEEEoverridecommandlockouts                              

\usepackage{amsmath} 
\usepackage{amssymb}  
\usepackage{cite}
\usepackage{amsfonts}
\usepackage{graphicx}
\usepackage{subfigure}
\usepackage[colorlinks,linkcolor=black,anchorcolor=black, citecolor=black, urlcolor=black]{hyperref}
\usepackage{multirow}
\usepackage{color}

\title{\LARGE \bf
RGB-D Grasp Detection via Depth Guided Learning \\ with Cross-modal Attention 
}

\author{Ran Qin, Haoxiang Ma, Boyang Gao and Di Huang$^{*}$
\thanks{This work is partly supported by the National Natural Science Foundation of China (62022011), the Research Program of State Key Laboratory of Software Development Environment (SKLSDE-2021ZX-04), and the Fundamental Research Funds for the Central Universities.}
\thanks{Ran Qin, Haoxiang Ma and Di Huang are with the State Key Laboratory of Software Development Environment, School of Computer Science and Engineering, Beihang University, Beijing, China.}
\thanks{Boyang Gao is with the Geometry Robotics and the School of Computer Science and Technology, Harbin Institute of Technology, Harbin, China.}
\thanks{*Corresponding author (email:  \href{mailto:dhuang@buaa.edu.cn}{dhuang@buaa.edu.cn}).}
}

\begin{document}

\maketitle
\thispagestyle{empty}
\pagestyle{empty}

\begin{abstract}

Planar grasp detection is one of the most fundamental tasks to robotic manipulation, and the recent progress of consumer-grade RGB-D sensors enables delivering more comprehensive features from both the texture and shape modalities. However, depth maps are generally of a relatively lower quality with much stronger noise compared to RGB images, making it challenging to acquire grasp depth and fuse multi-modal clues. To address the two issues, this paper proposes a novel learning based approach to RGB-D grasp detection, namely Depth Guided Cross-modal Attention Network (DGCAN). To better leverage the geometry information recorded in the depth channel, a complete 6-dimensional rectangle representation is adopted with the grasp depth dedicatedly considered in addition to those defined in the common 5-dimensional one. The prediction of the extra grasp depth substantially strengthens feature learning, thereby leading to more accurate results. Moreover, to reduce the negative impact caused by the discrepancy of data quality in two modalities, a Local Cross-modal Attention (LCA) module is designed, where the depth features are refined according to cross-modal relations and concatenated to the RGB ones for more sufficient fusion. Extensive simulation and physical evaluations are conducted and the experimental results highlight the superiority of the proposed approach.


\end{abstract}

\vspace{-3pt}
\section{Introduction}
\vspace{-1pt}

Grasp detection, which aims to generate gripper configurations on objects, is essential to robotic manipulation in the real world, such as bin-picking and sorting. Early methods \cite{DBLP:conf/icra/BicchiK00, DBLP:conf/icra/BuchholzFWW13} tackle this task by fitting the input to the templates with annotated grasps or analyzing shape characteristics in certain feature spaces. However, these methods either only work on known objects or suffer from high computational complexity, making them less practical for widespread use. Recently, with the rapid development of deep learning, data-driven methods \cite{DBLP:journals/ral/ChuXV18, DBLP:conf/rss/MorrisonLC18, DBLP:conf/rss/MahlerLNLDLOG17, DBLP:journals/corr/abs-2212-05275, DBLP:conf/iros/WangZG021} have dominated this field and shown great potential to fulfill a diversity of applications.

As a typical scenario, planar grasp detection has received increasing attention during the past decade, where the gripper configuration is usually represented as a 5-dimensional rotated rectangle on the image \cite{DBLP:conf/icra/JiangMS11}, and advanced object detection networks are adapted for prediction. Considering rich texture information in the color space, a large number of studies have been made in the literature \cite{DBLP:conf/iros/ZhangLBZTZ19, DBLP:conf/iros/ZhouLZTZZ18,DBLP:conf/icra/GuoSLKFX17}, which take RGB images as input. Meanwhile, due to the innovation of depth sensors, a trend has appeared that object shapes are introduced to complement appearances \cite{DBLP:conf/iros/KumraJS20, DBLP:journals/ral/ChuXV18}.

\begin{figure}[t]
\setlength{\abovecaptionskip}{0pt}
\centering
\subfigure[Measured grasp depth]{  
\includegraphics[width=0.45\linewidth]
{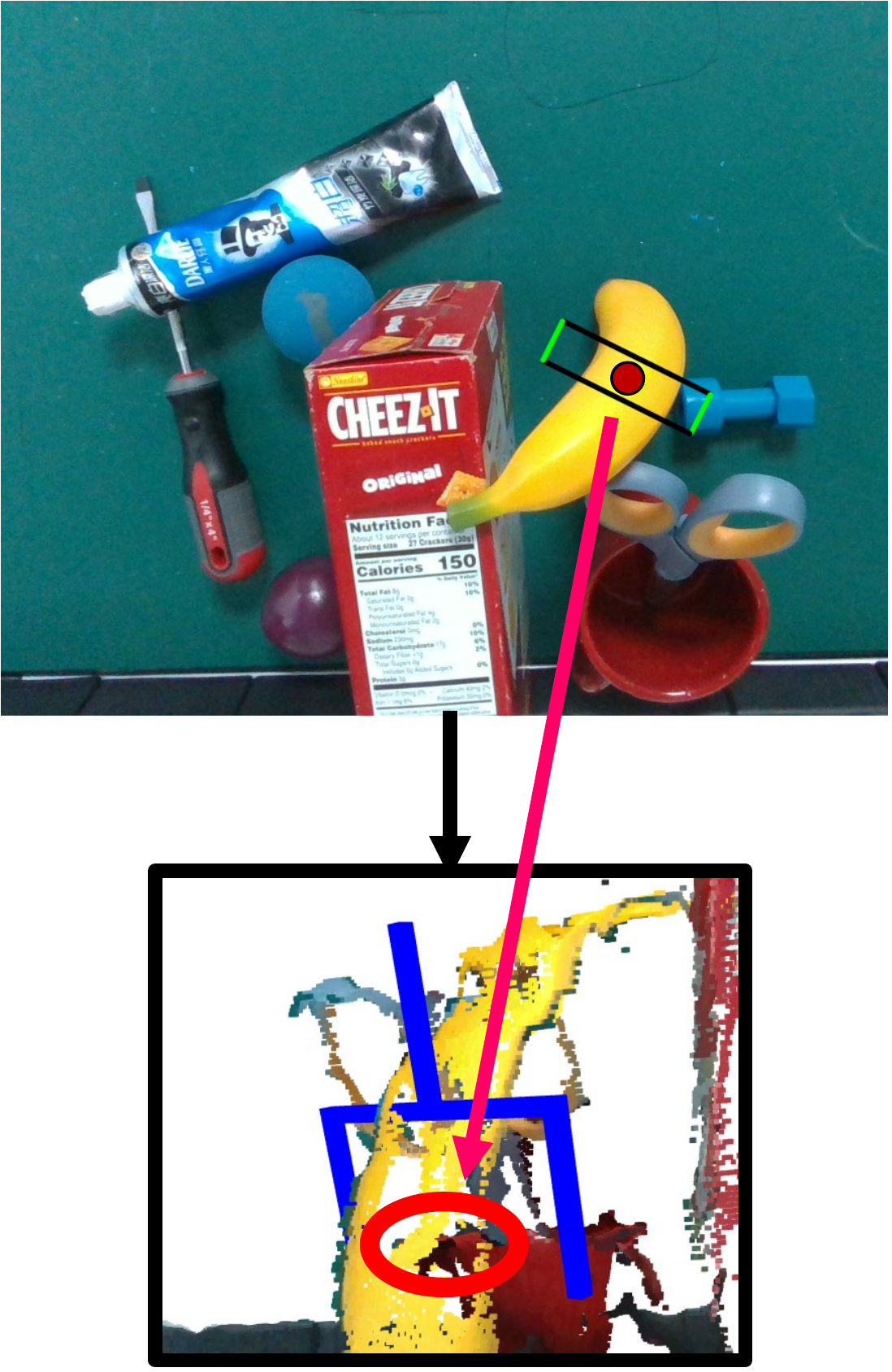}
}
\subfigure[Predicted grasp depth]{ 
\includegraphics[width=0.45\linewidth]{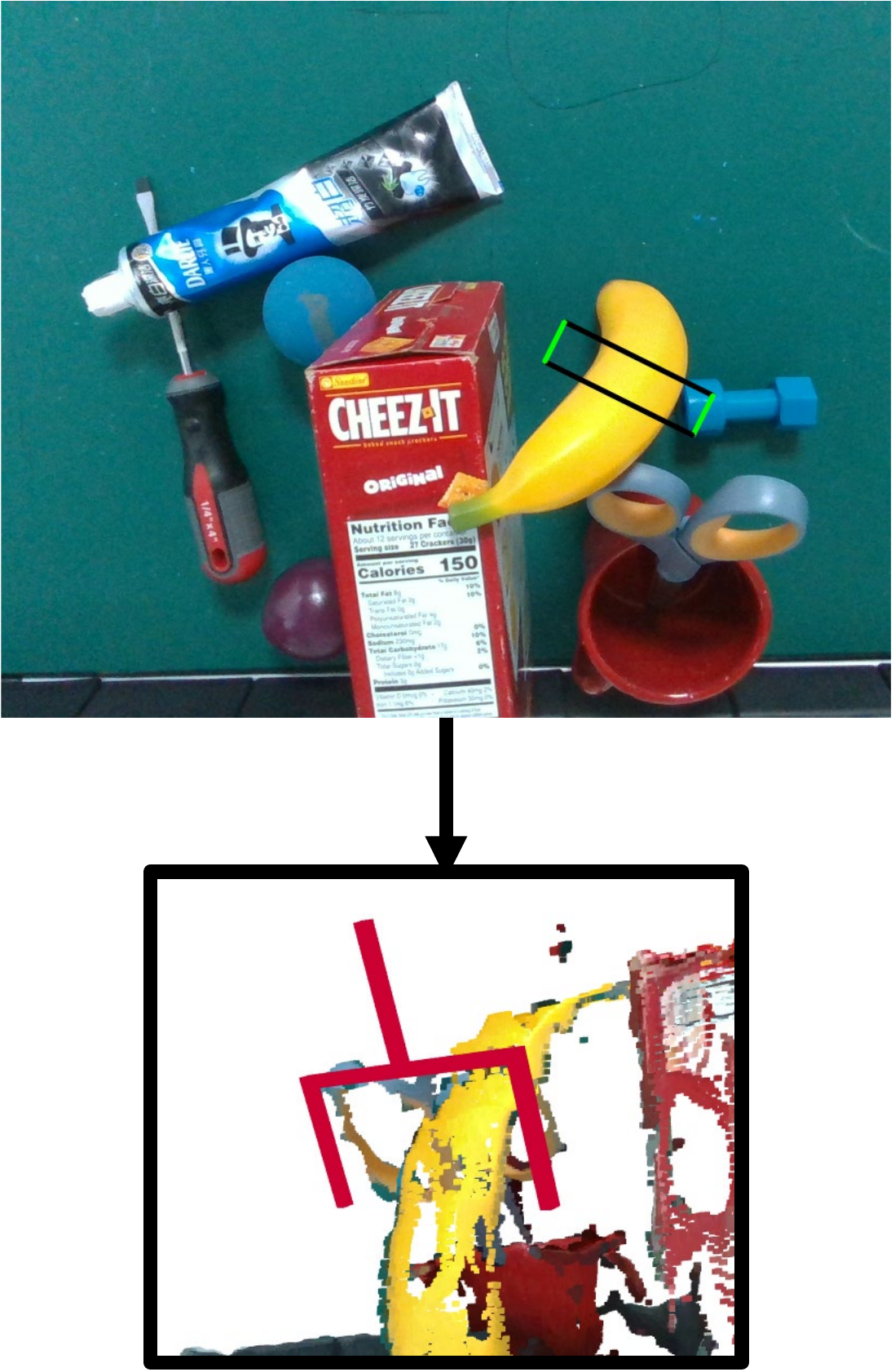}
}
\caption{Visualization of detected grasps which make use of depth clues in two different ways. The color of the grasp represents the score in terms of the force closure metric. The red indicates a grasp of a high quality while the blue denotes a grasp of a low quality. (a) shows the grasp which adopts the measured depth in the grasp center and (b) shows the grasp with the predicted depth achieved by the proposed method.}
\label{visualization about depth prediction}
\vspace{-20pt}
\end{figure}

Despite promising performance reported in RGB-D grasp detection \cite{DBLP:conf/iros/KumraK17, DBLP:journals/ral/ChuXV18, DBLP:conf/iros/KumraJS20}, two major issues still remain challenging, because for mainstream sensing devices, \emph{e.g.} Intel RealSense and Microsoft Kinect, the data in the depth channel is of a relatively lower quality with much stronger noise compared to that in the RGB channel. One is how to effectively and efficiently acquire the grasp depth, and another is how to sufficiently combine the credits in the RGB and depth modalities.

For the former issue, a major manner is to directly employ the depth value of the raw data in the center of the predicted 5-dimensional rectangle \cite{DBLP:conf/icra/JiangMS11, DBLP:journals/ijrr/MorrisonCL20}. While efficient, the depth value is often not so accurate or even missing in such noisy images, which is prone to incur failed grasps colliding with objects, as shown in Fig. \ref{visualization about depth prediction} (a). An alternative is to initially sample a set of patches and then estimate grasp depths from them \cite{DBLP:conf/rss/MahlerLNLDLOG17, DBLP:journals/ral/SatishMG19, DBLP:conf/iros/GariepyRCG19, DBLP:journals/mva/WangLHFSQ20}. These methods indeed boost the accuracy to some extent, but the entire procedure is complicated and incurs huge computational cost, especially in cluttered scenes. Both the two strategies make the generalization to the real-world case problematic. 
For the latter issue, existing methods conduct the fusion of RGB and depth data by summing or concatenating at the early or middle stage of the networks \cite{DBLP:conf/icra/RedmonA15, DBLP:journals/ral/ChuXV18, DBLP:conf/iros/KumraK17, DBLP:conf/icra/ZengSYDHBMTLRFA18}, where the two modalities are analogously treated.
This straightforward way confirms the complementarity of the two types of inputs; unfortunately, it does not fully take their properties into account. In particular, the depth images are generally coarse, and the extracted depth features tend to be misleading and thus bring difficulties in multi-modal clue integration. This dilemma is more serious when the networks become deeper \cite{DBLP:conf/cvpr/JiLYZPYB0ZL021, DBLP:conf/eccv/Chen020}. To the best of our knowledge, these two issues have not been well discussed and leave much space for improvement.

To address the issues aforementioned, in this paper, we propose a novel two-stage approach, namely Depth Guided Cross-modal Attention Network (DGCAN), to RGB-D grasp detection. Specifically, 
DGCAN builds a depth guided learning framework, where both the RGB and depth images are fed and their features are combined to generate grasp proposals. To better leverage the geometry clues in the noisy depth map, a complete 6-dimensional rectangle representation is adopted with the depth of the grasp dedicatedly considered besides its center, size, and orientation defined in the common 5-dimensional one. The prediction of the extra grasp depth substantially strengthens grasp features, thereby contributing to more accurate results. Furthermore, a Local Cross-modal Attention (LCA) module is designed to fuse the features which are separately encoded in the RGB and depth channels by two sub-networks. To reduce the negative impact caused by the discrepancy of data quality in two modalities, LCA conducts in an asymmetric way that depth features are refined according to cross-modal relations and concatenated to the RGB ones. In addition, a comprehensive RGB-D planar grasp dataset is produced based on GraspNet-1Billion \cite{DBLP:conf/cvpr/FangWGL20} to facilitate this research. Extensive simulation and real-world evaluations are made and the experimental results highlight the superiority of the proposed approach.

\vspace{-2pt}
\section{Related Work}


\vspace{-2pt}
\subsection{RGB Grasp Detection}
\vspace{-1pt}

In the past decade, many efforts have been made for grasp detection on RGB images, and we can see its development from hand-crafted methods to learning based ones. Following a sample-based pipeline, \cite{DBLP:journals/ijrr/SaxenaDN08} computes the edge and texture filter responses to decide whether an image patch contains a grasp point. Further, \cite{DBLP:conf/icra/PintoG16} extracts deep features by AlexNet \cite{DBLP:conf/nips/KrizhevskySH12} to estimate the graspable probabilities of different orientations. With the inspiration of the two-stage object detectors, such as Faster-RCNN \cite{DBLP:conf/nips/RenHGS15} and RRPN \cite{DBLP:journals/tmm/MaSYWWZX18}, end-to-end trainable networks \cite{DBLP:conf/icra/DepierreD021, DBLP:conf/iros/ZhouLZTZZ18, DBLP:conf/icra/AinetterF21} are proposed to directly locate rotated grasp rectangles with consistently improved performance. By applying the backbones pretrained on ImageNet \cite{DBLP:conf/nips/KrizhevskySH12}, they also alleviate the over-fitting risk, even with limited grasp annotations. Although RGB grasp detection methods achieve a great success, they are still not qualified enough for real-world applications due to the intrinsic ambiguity caused by 2D imaging techniques.

\vspace{-3pt}
\subsection{RGB-D Grasp Detection}
\vspace{-2pt}

Depth data convey geometry clues, which are necessary to complement appearance ones to build more accurate features for grasp detection. On the one hand, to acquire grasp depth, some attempts investigate the simple and direct strategy which uses the value of the depth image in the center of the predicted rectangle \cite{DBLP:conf/icra/JiangMS11, DBLP:journals/ijrr/MorrisonCL20}, but the performance is sensitive to the quality of depth data, largely limited by current consumer-grade RGB-D sensors. Although some subsequent methods improve this by a sample-before-evaluate pipeline \cite{DBLP:conf/rss/MahlerLNLDLOG17, DBLP:journals/ral/SatishMG19, DBLP:conf/iros/GariepyRCG19, DBLP:journals/mva/WangLHFSQ20}, they suffer from high computational complexity, not easy to generalize to cluttered scenes. On the other hand, to jointly capture textural and geometric information, these methods conduct multi-modal combination in two major ways, \emph{i.e.} early fusion and middle fusion. For early fusion, RGB and depth images are stacked before feeding to the network. \cite{DBLP:conf/icra/JiangMS11} concatenates an RGB image and an aligned depth image as a 4-channel input. \cite{DBLP:conf/icra/RedmonA15, DBLP:journals/ral/ChuXV18} replace the blue channel in the RGB image with the depth channel to deliver a 3-channel matrix. However, as the distribution gap exists between modalities, it is difficult for a single network to achieve reconciliation. Regarding middle fusion, RGB and depth images are individually processed through two different networks with similar architectures and finally aggregated for prediction \cite{DBLP:conf/iros/KumraK17, DBLP:conf/icra/ZengSYDHBMTLRFA18, DBLP:conf/iros/ZhuLBCL0TTL20, song2022deep}. These methods perform better than early fusion based ones, but they generally treat the two modalities in an analogous manner and ignore their quality difference, leading to sub-optimal results.

\begin{figure}[t]
\setlength{\abovecaptionskip}{0pt}
\centering
\subfigure[]{  
\centering  
\includegraphics[width=0.3\linewidth]{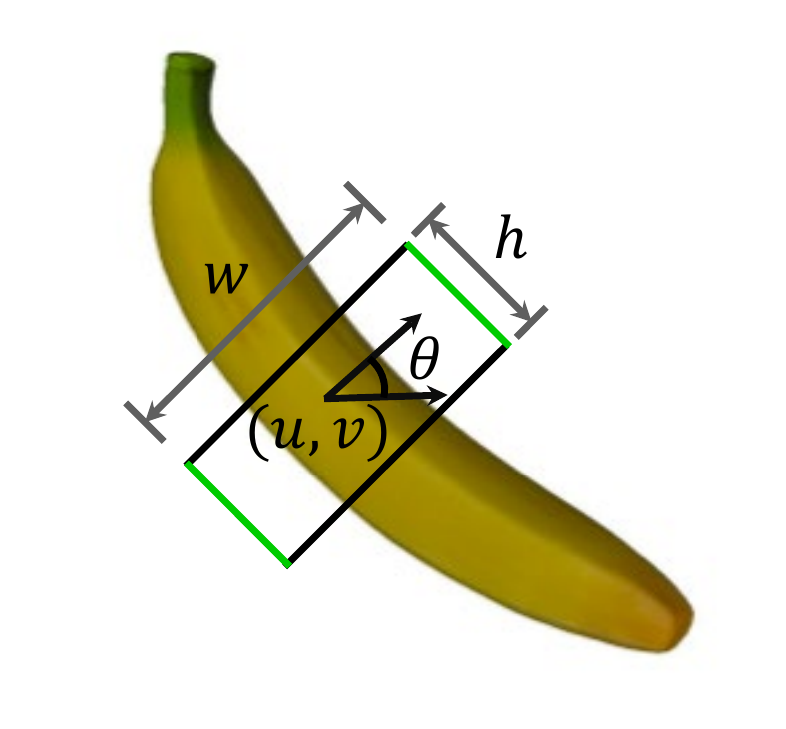}
}
\subfigure[]{ 
\centering    
\includegraphics[width=0.3\linewidth]{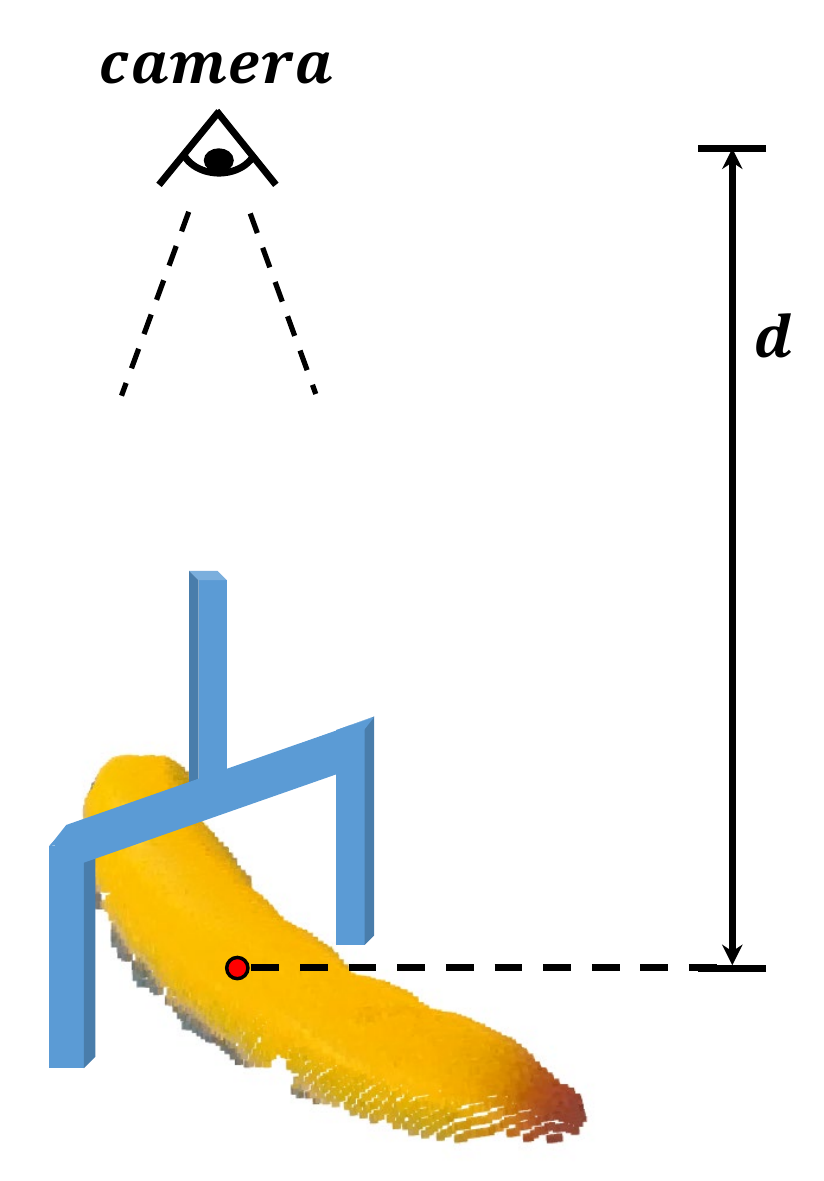}
}
\caption{(a) The 5-dimensional grasp configuration $(u, v, w, h, \theta)$. (b) The grasp depth $d$ in the 6-dimensional grasp configuration.}
\label{6d grasp configuration}
\vspace{-18pt}
\end{figure}

\begin{figure*}[ht]
\vspace{5pt}
\setlength{\abovecaptionskip}{0pt}
\centering
\subfigure[]{  
\centering  
\includegraphics[width=0.60\linewidth]{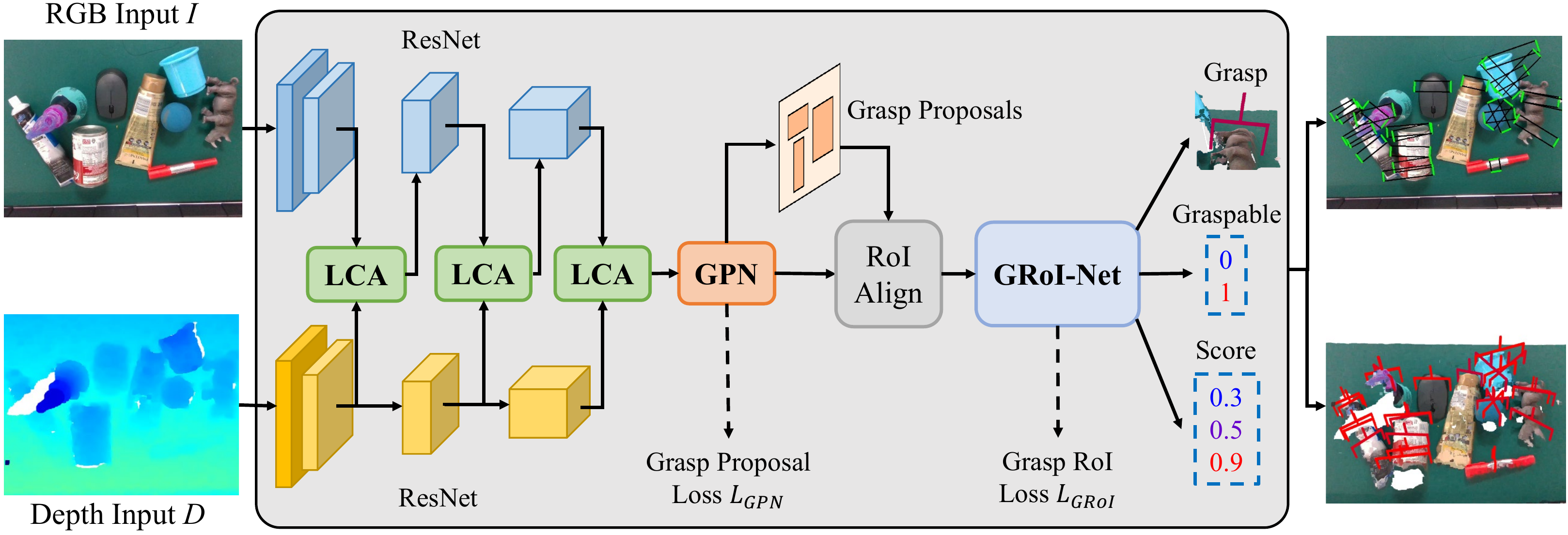}
\label{overall framework}}
\subfigure[]{  
\centering
\includegraphics[width=0.35\linewidth]{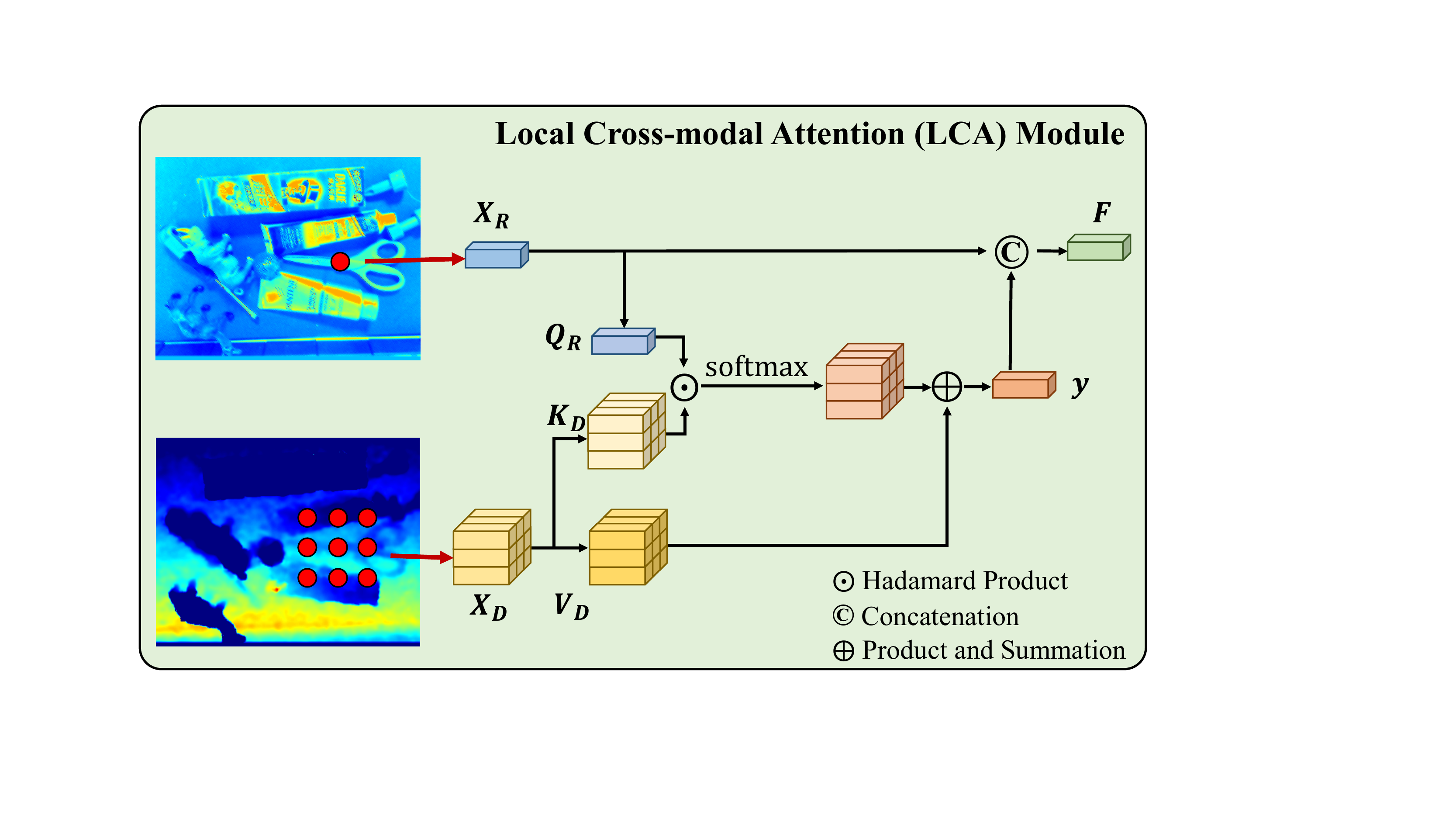}
\label{local cross-modal attention}
}
\caption{(a) Overview of the proposed DGCAN framework. It consists of three main parts: Local Cross-modal Attention (LCA) based multi-modal fusion network, Grasp Proposal Network (GPN) and Grasp Region of Interest Network (GRoI-Net) with depth prediction. (b) Details of the LCA module. RGB features are treated as queries and depth features are treated as keys and values.}
\label{DGCAN and LCA}
\vspace{-18pt}
\end{figure*}

\vspace{-7pt}
\section{Problem Formulation}


Given an RGB image $I \in \mathbb{R}^{3 \times H \times W}$ and its corresponding depth image $D \in \mathbb{R}^{1 \times H \times W}$ as input, different from the common 5-dimensional rectangle representation for planar grasp detection proposed in \cite{DBLP:conf/icra/JiangMS11}, we highlight the necessity of grasp depth to RGB-D grasp detection and adopt a 6-dimensional rectangle with an additional variable for grasp depth, as Fig. \ref{6d grasp configuration} shows, which is formally defined as:

\begin{equation}
\setlength{\abovedisplayshortskip}{-10pt}
\setlength{\belowdisplayshortskip}{2pt}
\textbf{g} = (u, v, d, w, h, \theta)
\label{6-dimensional grasp configurations equation}
\end{equation}
where $(u, v)$, $w$, $h$ and $\theta$ are the same as those defined in the 5-dimensional representation, denoting the center coordinate, width, height, and angle between the width and $x$-axis in the image of the grasp rectangle respectively. $d$ is  added to indicate the depth of the fingertips relative to the camera.




\vspace{-2pt}
\section{Methodology}
\vspace{-1pt}

\subsection{Overall Framework}
\vspace{-2pt}

With the 6-dimensional rectangle representation, DGCAN builds a depth guided learning framework. It follows the two-stage pipeline \cite{DBLP:journals/ral/ChuXV18} but extends it by multi-modal fusion and grasp depth prediction. As in Fig. \ref{DGCAN and LCA} (a), DGCAN is composed of three parts: \emph{i.e.} Grasp Proposal Network (GPN), Grasp Region of Interest Network (GRoI-Net), and Local Cross-modal Attention (LCA) based fusion network.



Taking a 3-channel RGB image and a 1-channel depth image as input, two networks with the same architecture (\emph{i.e.} ResNet \cite{DBLP:conf/cvpr/HeZRS16}) are adopted to extract appearance and geometry features respectively, where the depth image is replicated to 3 channels before processing. To sufficiently utilize the RGB and depth information, the LCA module is designed and placed after each convolutional block of the network, which performs an asymmetric fusion by first refining the depth features through cross-modal relation learning and then concatenating them to the RGB ones. The multi-modal features in the last block are further fed into GPN to generate possible grasp regions and predict 6-dimensional grasps in GRoI-Net. In the inference stage, we apply Non-Maximum Suppression (NMS) after GPN to suppress duplicated grasp proposals and retain final grasps according to the graspablility scores.


\subsection{Grasp Proposal Network}\label{Grasp Region Proposal Network}

As in \cite{DBLP:journals/ral/ChuXV18}, GPN is introduced as the first stage to estimate potential grasp candidates over the whole image. Concretely, given the multi-modal feature maps as input, it predicts the grasp bounding box $\hat{g} = (\hat{u}, \hat{v}, \hat{w}, \hat{h})$ and its probability of grasp proposal $\hat{p}$ for each anchor.



Compared to the typical Region Proposal Network (RPN) \cite{DBLP:conf/nips/RenHGS15} applied in the two-stage object detectors, GPN aims to find the optimal match from a dense collection of positive grasp labels with different scores. In this case, besides setting an Intersection over Union (IoU) threshold to filter Ground Truths (GTs) by positions and scales, only the grasp with the highest score $s$ is assigned to the anchor. The loss of GPN is defined as follows:



\begin{equation}
\begin{aligned}
    L_{GPN}(\{(\hat{p}_i, \hat{t}_i)\}_{i=1}^{N}) = \frac{1}{N_{cls}}\sum_i{L_{GPN}^{cls}(\hat{p}_i, \hat{p}_i^*)} 
    \\+ \lambda_1\frac{1}{N_{reg}}\sum_i{\hat{p}_i^*L_{GPN}^{reg}}(\hat{t}_i, \hat{t}_i^*)
\label{GPN loss}
\end{aligned}
\end{equation}
where $L_{GPN}^{cls}$ is the cross entropy loss of binary classification estimating whether an anchor is graspable, $L_{GPN}^{reg}$ is the Smooth L1 loss for grasp proposal regression, and $\lambda_1$ is a weight parameter. $N_{cls}$ and $N_{reg}$ indicate the number of sampled anchors and number of positive samples respectively. $\hat{p}_i$ and $\hat{p}_i^*$ denote the predicted probability and classification label of the $i$-th anchor. $\hat{p}_i^* = 1$ means a specific grasp and $\hat{p}_i^* = 0$ indicates that there is no grasp. $\hat{t}_i$ and $\hat{t}_i^*$ represent the predicted and GT of the 4-dimensional grasp vector. The predicted grasp proposal $(\hat{u}, \hat{v}, \hat{w}, \hat{h})$ is calculated as:

\begin{equation}
\begin{aligned}
    \hat{t}_{u} = (\hat{u} - u_a) / w_a, \quad&\hat{t}_{v} = (\hat{v} - v_a) / h_a,
    \\\hat{t}_{w} = \log(\hat{w} / w_a), \quad&\hat{t}_{h} = \log(\hat{h} / h_a)
\end{aligned}
\end{equation}
where $(u_a, v_a, w_a, h_a)$ denotes the grasp vector of an anchor. Then, RoI Align is employed to extract features from grasp proposals and feed them into the subsequent GRoI-Net.

\begin{table*}[ht]
\setlength{\abovecaptionskip}{0pt}
\setlength{\belowcaptionskip}{-0pt}
\vspace{5pt}
\caption{Summary of the properties of public planar grasp datasets.}
\label{dataset comparison}
\centering
\begin{tabular}{c|c|c|c|c|c|c|c}
\hline
\multirow{2}{*}{\textbf{Dataset}} & \textbf{Grasp} & \multirow{2}{*}{\textbf{Modality}} & \textbf{Objects} & \textbf{Grasps} & \textbf{Num} & \textbf{Num} & \textbf{Num}\\
~ & \textbf{Depth} & ~ & \textbf{/Img} & \textbf{/Img} & \textbf{Imgs} & \textbf{Objects} & \textbf{Grasps}\\
\hline
Levine \textit{et al.} \cite{DBLP:journals/ijrr/LevinePKIQ18} & No & RGB-D & - & 1 & 800K & - & 800K\\
\hline
Pinto \textit{et al.} \cite{DBLP:conf/icra/PintoG16} & No & RGB-D & - & 1 & 50K & 150 & 50K \\
\hline
Cornell \cite{DBLP:conf/icra/JiangMS11} & No & RGB-D & 1 & $\sim$8 & 1035 & 240 & 8K\\
\hline
Mahler \textit{et al.} \cite{DBLP:conf/rss/MahlerLNLDLOG17} & Yes & D & 1 & 1 & 6.7M & 1500 & 6.7M\\
\hline
Jacquard \cite{DBLP:conf/iros/DepierreD018} & No & RGB-D & 1 & $\sim$20 & 54k & 11K & 1.1M\\
\hline
VMRD \cite{DBLP:conf/iros/ZhangLBZTZ19} & No & RGB & $\sim$3 & $\sim$20 & 4.7K & $\sim$200 & 100K\\
\hline
Multi-Object \cite{DBLP:journals/ral/ChuXV18} & No & RGB-D & $\sim$4 & $\sim$30 & 96 & - & 2904\\
\hline
REGRAD \cite{DBLP:journals/ral/ZhangYWZLDZ22} & No & RGB-D & 1$\sim$20 & 1.02K & 900K & 50K & 100M \\
\hline
GraspNet-1Billion \cite{DBLP:conf/cvpr/FangWGL20} & No & RGB-D & $\sim$10 & 3$\sim$9M & 97K & 88 & $\sim$1.2B \\
\hline
GraspNet-Planar & Yes & RGB-D & $\sim$10 & $\sim$30K & $\sim$11K & 88 & 169M\\
\hline
\end{tabular}
\vspace{-15pt}
\end{table*}

\subsection{Grasp Region of Interest Network}
For each RoI, GRoI-Net further extracts the feature on it using the fourth block $C_4$ in ResNet \cite{DBLP:conf/cvpr/HeZRS16} and predicts the 6-dimensional grasp configuration $\textbf{g} = (u, v, d, w, h, \theta)$ with its grasp possibility $q$ and grasp score $s$. Therefore, given $M$ grasp proposals, the outputs of GRoI-Net are $M \times 7$, $M \times 2$ and $M \times 1$ for grasp target regression $t = (t_u, t_v, t_d, t_w, t_h, t_{sin}, t_{cos})$, binary classification $q$ and grasp score estimation $s$, respectively. Particularly, for grasp depth prediction, we consider the depth measurement $d_o$ captured by the RGB-D camera as the reference depth and regress the offset $t_d$ defined as:
\begin{equation}
    t_d = d - d_o.
\end{equation}
Meanwhile, we encode the angle as two components $t_{sin} = \sin(2\theta)$ and $t_{cos} = \cos(2\theta)$ in the range $[-1, 1]$ and calculate the grasp orientation as:
\begin{equation}
    \theta = \frac{1}{2}\arctan\frac{\sin(2\theta)}{\cos(2\theta)}.
\end{equation}
With this definition, the target function of GRoI-Net is defined as follows:
\begin{equation}
\begin{aligned}
    L_{GRoI}(\{(q_i, t_i, s_i)\}_{i=1}^{M}) = \frac{1}{M_{cls}}\sum_i{L_{GRoI}^{cls}(q_i, q_i^*)} 
    \\+ \lambda_2\frac{1}{M_{reg}}\sum_i{L_{GRoI}^{reg}(t_i, t_i^*)} 
    \\+ \lambda_3\frac{1}{M_{reg}}\sum_i{L_{GRoI}^{reg}(s_i, s_i^*)}
\end{aligned}
\end{equation}
where $q_i^*$, $t_i^*$ and $s_i^*$ denote the GTs for $q_i$, $t_i$ and $s_i$. $M_{cls}$ and $M_{reg}$ are similar to $N_{cls}$ and $N_{reg}$ in Eq. \ref{GPN loss}. Here, the cross entropy loss is used for $L_{GRoI}^{cls}$, while the Smooth L1 loss is used for $L_{GRoI}^{reg}$. $\lambda_2$ and $\lambda_3$ are weight parameters.

\subsection{Local Cross-modal Attention based Fusion Network}
The previous RGB-D grasp detection networks \cite{DBLP:conf/iros/KumraK17, DBLP:conf/icra/ZengSYDHBMTLRFA18, DBLP:conf/iros/ZhuLBCL0TTL20} generally integrate RGB and depth features by summation or concatenation with the two modalities analogously treated. However, the data in the depth channel is of a lower quality than that in the RGB channel in existing RGB-D sensors and these methods do not fully take this gap into account, thus leading to sub-optimal fusion. To handle this, we dedicatedly design a new module, \emph{i.e.} LCA, to combine the RGB and depth features in an asymmetric manner, where the relatively stable RGB features are used to refine the noisy depth ones. Furthermore, as local shape perception is critical to grasp detection, for the depth feature in each position, only the ones within a neighborhood are involved.
As illustrated in Fig. \ref{DGCAN and LCA} (b), LCA first refines depth features by weighting them according to the relation between the RGB and depth modalities in a local region and then concatenates the weighted depth features to those in the RGB stream.

Denote the RGB and depth feature as $X_{R} \in \mathbb{R}^{C \times H \times W}$ and $X_D \in \mathbb{R}^{C \times H \times W}$ respectively, the output of LCA $F$ is formulated as:
\begin{equation}
    F = f(X_{R} || r(X_{R}, X_D))
\end{equation}
where $||$ is the concatenation operation of the feature maps from the two modalities, $f$ refers to $1\times1$ convolution with the number of channels modified from $2C$ to $C$, and $r$ represents the depth refinement operator based on cross-modal attention. Concretely, $X_{R}$ serves as the \textit{query} and is embedded to $Q_{R} \in \mathbb{R}^{C' \times H \times W}$; $X_{D}$ serves as the \textit{key} and \textit{value}, embedded to $K_{D} \in \mathbb{R}^{C' \times H \times W}$ and $V_{D} \in \mathbb{R}^{C' \times H \times W}$ by two different $1\times1$ convolutions. Inspired by \cite{ramachandran2019stand}, we design an attention module to capture the relationship between the RGB feature and the corresponding depth features within a neighborhood considering the importance of local geometric features to grasp detection. Given an RGB query at $(i, j)$, we extract depth keys and values in the neighboring region $N_{k}(i, j)$ with spatial size $k$ and calculate the depth output $v_{ij}$ as follows:
\begin{equation}
    v_{ij} = \sum_{(m,n) \in N_{k}(i, j)}{\text{softmax}(Q_{R,ij}K_{D,mn} / \sqrt{C'})V_{D,mn}}
\end{equation}
where $(m, n) = (i + \Delta{i}, j + \Delta{j})$, $\Delta{i} = -\lceil k/2 \rceil, -\lceil k/2 \rceil + 1, ..., \lceil k/2 \rceil$, $\Delta{j} = -\lceil k/2 \rceil, -\lceil k/2 \rceil + 1, ..., \lceil k/2\rceil$. We then project the outputs $v$ to the final depth features $y \in \mathbb{R}^{C \times H \times W}$ by $1 \times 1$ convolution.
Following the self-attention \cite{DBLP:conf/nips/VaswaniSPUJGKP17} strategy, we implement the multi-head attention mechanism and add sine positional encodings in queries and keys.

\subsection{Loss Function}
During training, the whole network is optimized in an end-to-end manner by minimizing the following loss function:
\begin{equation}
    L = L_{GPN}(\{(\hat{p}_i, \hat{t}_i)\}_{i=1}^{N}) + L_{GRoI}(\{(q_i, t_i, s_i)\}_{i=1}^{M})
\end{equation}



\begin{table*}[ht]
\setlength{\abovecaptionskip}{0pt}
\setlength{\belowcaptionskip}{0pt}
\vspace{5pt}
\caption{Performance comparison on GraspNet-Planar captured by RealSense/Kinect. \textit{CD} represents collision detection.}
\label{comparision with others}
\centering
\resizebox{\textwidth}{!}{
\begin{tabular}{c|c c c|c c c|c c c}
\hline
\multirow{2}{*}{\textbf{Method}} & \multicolumn{3}{c|}{\textbf{Seen}} & \multicolumn{3}{c|}{\textbf{Similar}} & \multicolumn{3}{c}{\textbf{Novel}}\\
\cline{2-10}
~ & \textbf{AP} & \textbf{AP$_{0.8}$} & \textbf{AP$_{0.4}$} & \textbf{AP} & \textbf{AP$_{0.8}$} & \textbf{AP$_{0.4}$} & \textbf{AP} & \textbf{AP$_{0.8}$} & \textbf{AP$_{0.4}$} \\
\hline
GR-ConvNet \cite{DBLP:conf/iros/KumraJS20} & 19.97/17.72 & 25.42/22.23 & 11.43/10.60 & 13.24/13.04 & 16.93/16.39 & 6.79/7.67 & 6.32/4.41 & 7.67/5.41 & 1.69/1.52 \\
GG-CNN2 \cite{DBLP:journals/ijrr/MorrisonCL20} & 27.45/23.73 & 35.31/30.20 & 16.60/14.90 & 21.49/18.02 & 27.87/23.44 & 11.27/10.12 & 9.76/7.12 & 11.66/8.77 & 2.95/3.14\\
Chu \textit{et al.} \cite{DBLP:journals/ral/ChuXV18} & 29.71/25.69 & 35.98/30.70 & 23.76/20.96 & 24.17/20.92 & 30.00/25.44 & 16.86/15.87 & 10.92/6.70 & 13.38/8.23 & 6.69/3.88 \\
\hline
DGCAN & 49.85/47.32 & 59.67/57.27 & 42.24/38.55 & 41.46/35.73 & 50.31/44.22 & 33.69/26.99 & 17.48/16.10 & 21.83/20.01 & 7.90/7.81 \\
DGCAN-\textit{CD} & \textbf{52.16/50.45} & \textbf{62.71/61.22} & \textbf{43.14/40.64} & \textbf{44.69/38.62} & \textbf{54.52/47.85} & \textbf{35.37/28.81} & \textbf{19.26/17.66} & \textbf{23.93/21.94} & \textbf{8.89/8.29}\\
\hline
\end{tabular}}
\vspace{-10pt}
\end{table*}

\section{GraspNet-Planar Database}

Current public RGB-D planar grasp data do not contain grasp depth annotations, we thus build a new benchmark based on a 6-DoF grasp dataset, \emph{i.e.} GraspNet-1Billion \cite{DBLP:conf/cvpr/FangWGL20}, for evaluation. Following the same protocol as GraspNet-1Billion, we take 100 scenes for training and 90 scenes for testing, where the test set is divided into three parts for seen, similar and novel objects respectively. Because the camera pose of planar grasping is constrained to be perpendicular to the workspace plane from top to down, we only use the images where the angles between the camera's observation view and $z$-axis are smaller than $15^{\circ}$. In general, our benchmark consists of 5,513 RGB-D images from RealSense D435 and 5,341 images from Azure Kinect. 



Although GraspNet-1Billion provides transformation matrices from 6-DoF grasps to planar grasps, the effect of grasp depths is ignored, which incurs inaccurate grasp annotations. To deal with this, we present a new pipeline. Specifically, for each object in the scene, we first select grasps with small angles between the grasp approaches and camera view. The approach directions of the selected 6-DoF grasps are set the same as the camera view. This projection may result in failed grasp labels due to the change of grasp poses; therefore, we re-evaluate these grasps by collision detection as well as force closure metric \cite{DBLP:journals/ijrr/Nguyen88}. We assign new force-closure scores to the successful grasps as grasp scores and remove the failed samples. Finally, to obtain the gripper configurations of planar grasp detection, the grasp label $\textbf{G}$ in the $SE(3)$ space is projected to the image plane as follows:

\begin{equation}
    \textbf{G} = T_{CI}\textbf{g}
\label{transformation from image to camera}
\end{equation}
where $T_{CI}$ is the intrinsic parameter matrix of the camera and $\textbf{g}$ is described in Eq. \ref{6-dimensional grasp configurations equation}. Table \ref{dataset comparison} compares our GraspNet-Planar benchmark with major public counterparts. It provides not only multi-modal RGB-D images but also more complete annotations with grasp depths than others, which highlights its advantage for RGB-D planar grasp detection research.

\begin{figure}[b]
\setlength{\abovecaptionskip}{0pt}
\vspace{-15pt}
\centering
\subfigure[]{  
\centering  
\includegraphics[width=0.31\linewidth]{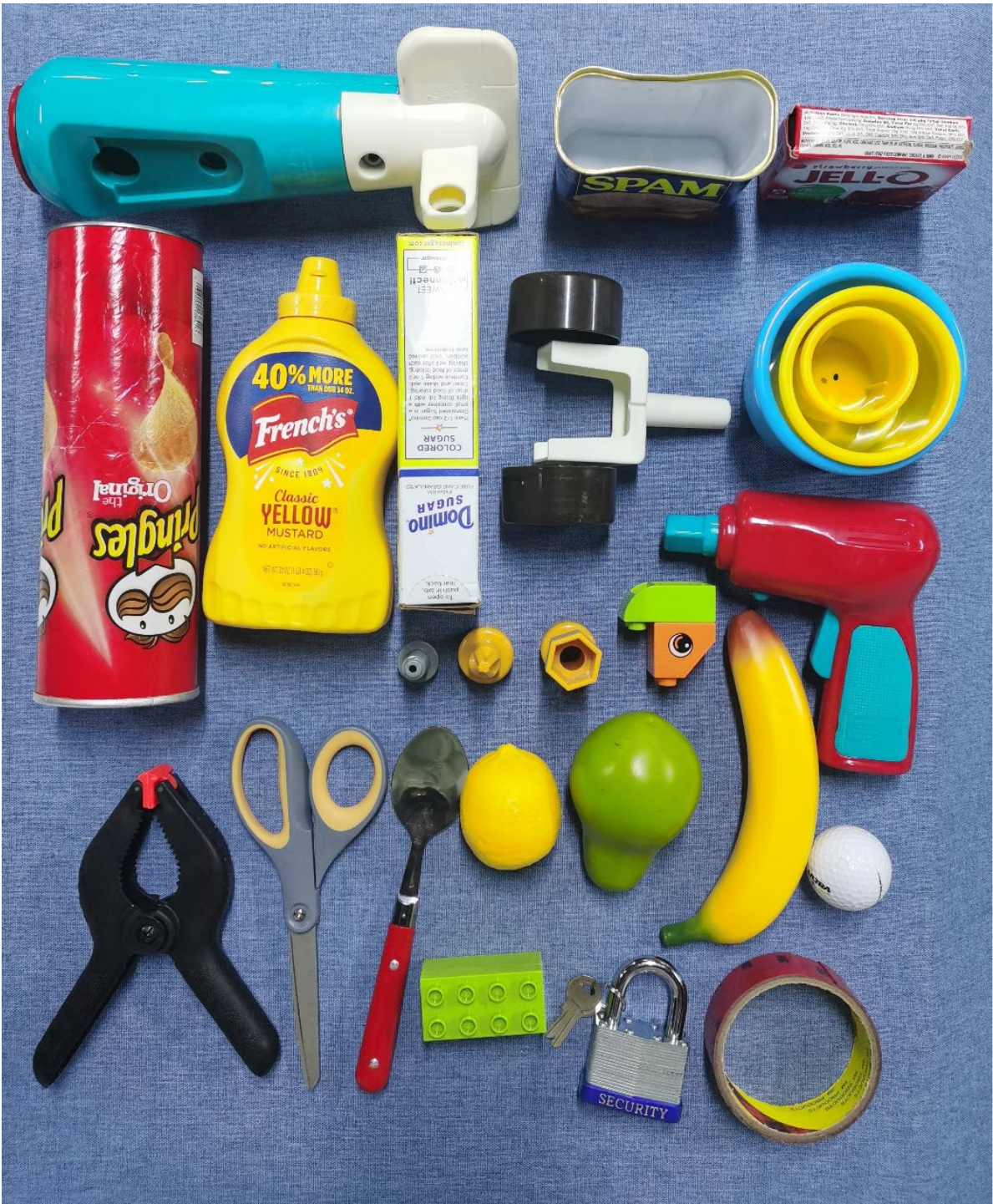}
\label{object set}
}
\subfigure[]{ 
\centering    
\includegraphics[width=0.5\linewidth]{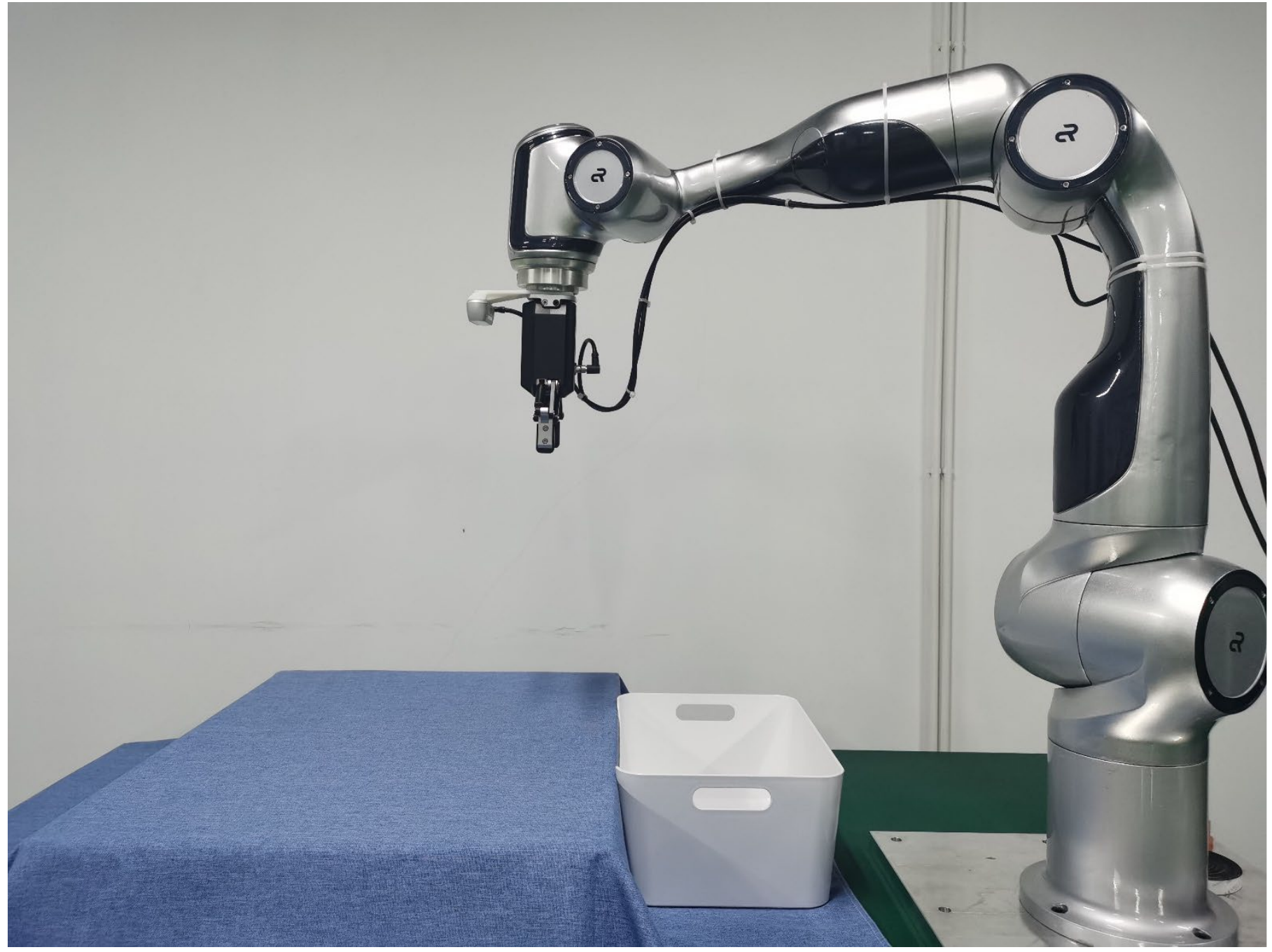}
\label{real-robot setting}
}
\caption{(a) The 25 household objects used in physical evaluation. (b) The hardware set-up with a 7-DoF Agile Diana-7 robot arm and an Intel RealSense D435i camera tied on the arm.}
\label{real-robot experiments}
\end{figure}


\section{Experiments}
\subsection{Protocols}

The simulation evaluation on GraspNet-Planar follows the metric in the GraspNet-1Billion benchmark \cite{DBLP:conf/cvpr/FangWGL20}, \emph{i.e.} Average Precision (\textbf{AP}). Given the predicted 6-dimensional planar grasps, we calculate $\textbf{AP}_{\mu}$ with different friction coefficient $\mu$ after grasp-NMS \cite{DBLP:conf/cvpr/FangWGL20}. As in GraspNet-1Billion, the overall result \textbf{AP} is calculated by averaging of $\textbf{AP}_{\mu}$, where $\mu$ ranges from $0.2$ to $1.0$ with the interval $\Delta\mu=0.2$.




For physical evaluation, we choose 25 objects of various sizes, shapes and textures from the YCB Object Set \cite{DBLP:journals/ram/CalliWSSAD15}, as shown in Fig. \ref{real-robot experiments} (a). We conduct all real-world experiments on a 7-DoF robotic arm in two settings: single-object grasping and multi-object grasping. For single-object grasping, 25 objects are employed and each object is evaluated in three different poses. For multi-object grasping, we construct 10 cluttered scenes, each of which contains 5 different objects. Grasping methods are required to remove the objects in scenes as many as possible within 10 attempts. As for the metric, Grasp Success Rate (GSR) is used in both single-object grasping and multi-object grasping, and Scene Completion Rate (SCR) is additionally recorded for multi-object grasping.

\vspace{-2.5pt}
\subsection{Implementation Details}
\vspace{-0.5pt}

We build our network based on ResNet-50\cite{DBLP:conf/cvpr/HeZRS16}. For LCA in the three residual blocks, we adopt the local spatial size $k = (5, 3, 3)$, which is further discussed in Sec. \ref{ablation study section}. The anchor aspect ratios and scales in GPN are set to [0.5, 1, 2] and [32, 64, 128, 256], respectively.



During training, the model is initialized by the weights pre-trained on ImageNet \cite{DBLP:conf/nips/KrizhevskySH12}. We choose 512 samples for both GPN and GRoI-Net, where the ratios of positive and negative samples are respectively 1:1 and 1:3. After GPN, we take 2,000 proposals whose IoUs are lower than 0.7. The weights in the loss function are set as $\lambda_1$ = 1, $\lambda_2$ = 1 and $\lambda_3$ = 4. During inference, we retain the predicted grasps with the graspable scores higher than $0.5$ to deliver the final result.




The model is trained on four GTX1080Ti GPUs by the SGD optimizer with the momentum and weight decay set to 0.9 and 0.0001. The mini-batch size is 4 and the initial learning rate is $5 \times 10^{-4}$ and decays at 43k iterations. Before training, grasp-NMS \cite{DBLP:conf/cvpr/FangWGL20} is launched to filter out largely overlapped GTs. Random horizontal flipping and rotating are employed to augment input images. Besides, the hole-filling filter and bilateral filter are applied to depth images for smoothing.



In physical evaluation, we use a 7-DoF Agile Diana-7 robot arm with an Intel RealSense D435i camera tied on it. The RGB-D images are captured above the workspace, as shown in Fig. \ref{real-robot experiments} (b). Our network and control algorithm are equipped on a desktop with an AMD Ryzen 5 2600 six-core processor and a single NVIDIA GeForce 1080 GPU.



\begin{figure}[!t]
\setlength{\abovecaptionskip}{0pt}
\centering  
\includegraphics[width=\linewidth]{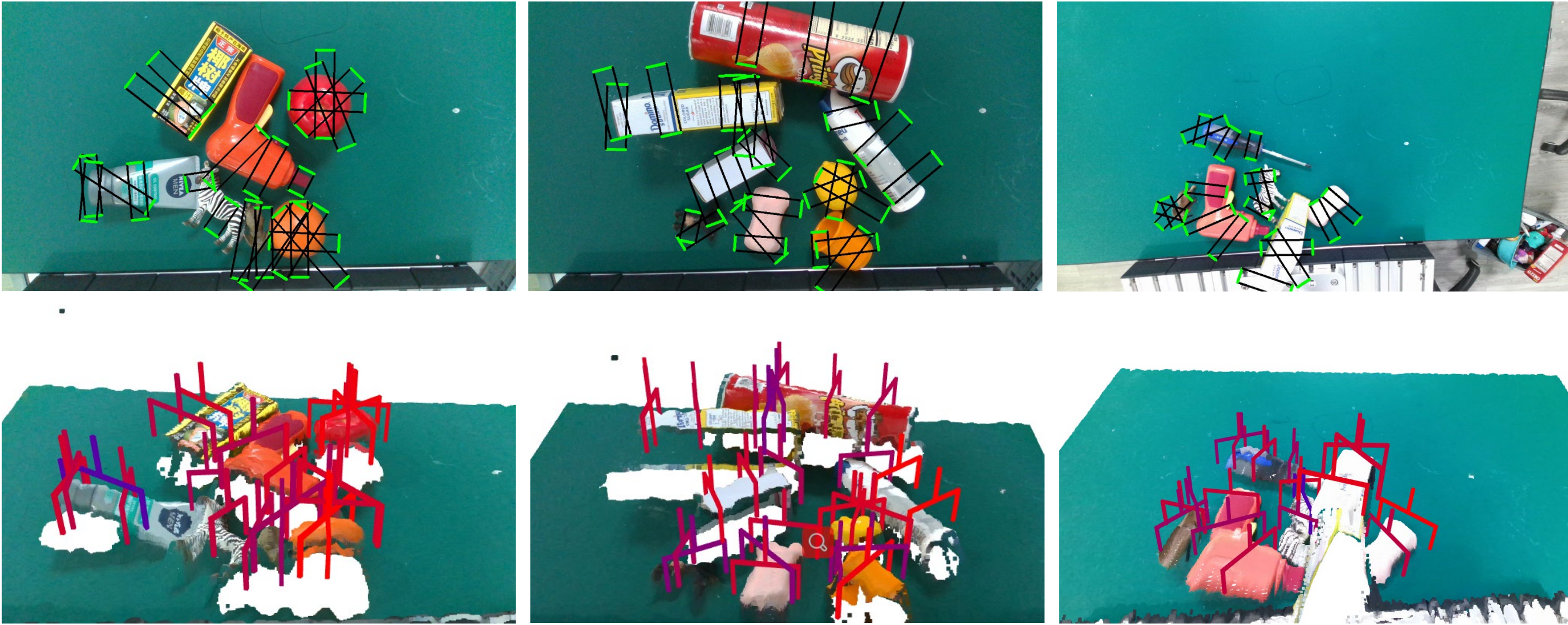}
\caption{Qualitative visualization on GraspNet-Planar captured by RealSense (the first 2 columns) and Kinect (the last column) after grasp-NMS \cite{DBLP:conf/cvpr/FangWGL20}. The color in the second row denotes the predicted scores of grasp poses ranging from 0 to 1. The blue color represents 0 and the red color represents 1. Zoom for better views.}
\label{result visualization}
\vspace{-19pt}
\end{figure}

\vspace{-3pt}
\subsection{Simulation Evaluation}
\vspace{-0.5pt}

Table \ref{comparision with others} summarizes the results of different solutions on the GraspNet-Planar dataset. We compare the proposed approach to three representative end-to-end trainable planar grasp detectors \cite{DBLP:journals/ral/ChuXV18, DBLP:journals/ijrr/MorrisonCL20, DBLP:conf/iros/KumraJS20}. \cite{DBLP:journals/ral/ChuXV18} is a two-stage network which is built based on Faster-RCNN \cite{DBLP:conf/nips/RenHGS15} and \cite{DBLP:journals/ijrr/MorrisonCL20, DBLP:conf/iros/KumraJS20} follow the one-stage pipeline which output dense predictions of grasp qualities, widths and orientations. It can be seen that our network outperforms the other state-of-the-art counterparts by a large margin, which demonstrates the effectiveness of grasp depth prediction and the LCA module. Some qualitative results are shown in Fig. \ref{result visualization}, which indicate the validity of our approach in cluttered scenes.

\begin{table}[h]
\vspace{5pt}
\setlength{\abovecaptionskip}{-5pt}
\setlength{\belowcaptionskip}{-0pt}
\caption{Physical robot evaluation results}
\label{physical robot experiment}
\begin{center}
\resizebox{\linewidth}{!}{\begin{tabular}{c|c|c|c}
\hline
\multirow{2}{*}{\textbf{Method}} & \textbf{Single-Object} & \multicolumn{2}{c}{\textbf{Multi-Object}}\\
\cline{2-4}
~ & \textbf{GSR (\%)} & \textbf{GSR (\%)} & \textbf{SCR (\%)} \\
\hline
Dex-Net 4.0\cite{DBLP:journals/scirobotics/MatlSDDMG19} & 70.67 (53/75) & 67.14 (47/70) & 94.00 (47/50) \\
\hline
FC-GQ-CNN\cite{DBLP:journals/ral/SatishMG19} & 73.33 (55/75) & 67.61 (48/71) & 96.00 (48/50)\\
\hline
DGCAN & \textbf{88.00} (66/75) & \textbf{84.75} (50/59) & \textbf{100.00} (50/50)\\
\hline
\end{tabular}}
\end{center}
\vspace{-10pt}
\end{table}

\begin{table}[h]
\setlength{\abovecaptionskip}{0pt}
\setlength{\belowcaptionskip}{-0pt}
\caption{Ablation study of grasp depth prediction on GraspNet-Planar captured by RealSense.}
\label{ablation study on grasp depth}
\centering
\begin{tabular}{c|c|c|c}
\hline
{\textbf{Grasp Depth}} & {\textbf{Seen}} & {\textbf{Similar}} & {\textbf{Novel}}\\
\hline
Center  & 42.01 & 34.54 & 15.25\\
\hline
Classification  & 48.57 & 39.01 & 17.06\\
\hline
Ours & \textbf{49.85} & \textbf{41.46} & \textbf{17.48}\\
\hline
\end{tabular}
\vspace{-15pt}
\end{table}

\begin{table}[h]
\vspace{5pt}
\setlength{\abovecaptionskip}{0pt}
\setlength{\belowcaptionskip}{-0pt}
\caption{Ablation study of multi-modal fusion mechanism on GraspNet-Planar captured by RealSense.}
\label{ablation study on multi-modal fusion}
\centering
\begin{tabular}{c|c|c|c}
\hline
{\textbf{Fusion Method}} & {\textbf{Seen}} & {\textbf{Similar}} & {\textbf{Novel}}\\
\hline
RGB & 40.83 & 27.76 & 9.94\\
\hline
RGD & 47.87 & 38.05 & 16.07\\
\hline
RGB-D-Sum & 48.08 & 38.95 & 15.43\\
\hline
RGB-D-Concat & 48.61 & 40.46 & 16.82\\
\hline
LCA & \textbf{49.85} & \textbf{41.46} & \textbf{17.48}\\
\hline
\end{tabular}
\end{table}

\subsection{Physical Evaluation}

In real-world experiments, Dex-Net 4.0 \cite{DBLP:journals/scirobotics/MatlSDDMG19} and FC-GQ-CNN \cite{DBLP:journals/ral/SatishMG19} which also involve the policies of acquiring grasp depths are employed for comparison. Dex-Net 4.0 takes measurements of grasp centers in depth images as grasp depths and FC-GQ-CNN uniformly samples depth values to several bins within the range of the whole scene and then evaluates them. As shown in Table \ref{physical robot experiment}, Dex-Net 4.0 and FC-GQ-CNN achieve similar performance which fall behind ours in both settings.

\subsection{Ablation Study} \label{ablation study section}



\textbf{Influence of depth guided learning.} We use two baselines, both of which deliver grasp depths. One is \textit{Center}, where we predict 5-dimensional grasp rectangles and set the sum of the depth measurements of the rectangle centers and a fixed offset of 20 millimeters as the grasp depths. Another is \textit{Classification}. Here, similar to \cite{DBLP:journals/ral/SatishMG19}, the prediction of grasp depths is modeled as a classification problem. We replace the regression head of grasp depths with a classification head. The grasp depths are uniformly divided into 40 bins from the maximum to the minimum depth value of scenes. The width of each bin is about 6 millimeters. It should be noted that for \textit{Classification} we train the networks with different numbers of depth bins ranging from 25 to 50 by an interval of 5 and the one with 40 bins achieves the best performance.
For fair comparison, we adopt very similar network architectures in all the models, which only differ in the head of GRoI-Net.

As shown in Table \ref{ablation study on grasp depth}, the model that acquires grasp depths by \textit{Center} reaches the lowest performance among the three, which indicates the significance of replacing the measured grasp depth to the predicted depth. When compared to \textit{Classification}, the proposed approach has respectively gained $1.28\%$, $2.45\%$ and $0.42\%$ improvements on seen, similar and novel objects, proving that the grasp depth predicted by regression is more accurate than that by classification. Besides, \textit{Classification} brings extra hyper-parameters, which increases the design complexity, such as the range of depth values and the number of depth bins.

\begin{figure}[htbp]
\setlength{\abovecaptionskip}{-5pt}
\setlength{\belowcaptionskip}{0pt}
\vspace{-5pt}
\centering
\includegraphics[width=0.85\linewidth]{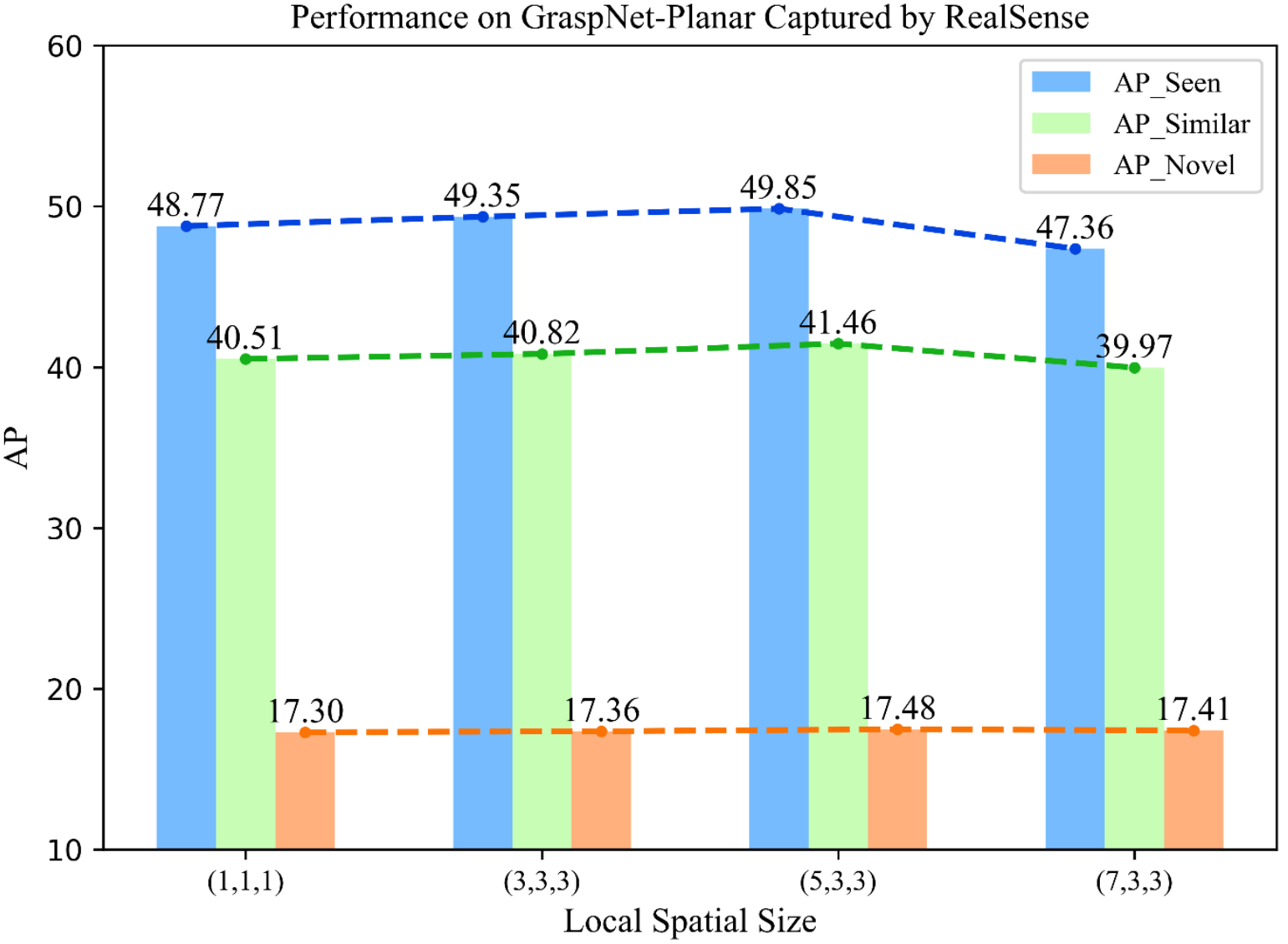}
\caption{AP under various local spatial sizes $(k_1, k_2, k_3)$ for different stages $(C_1, C_2, C_3)$ of ResNet.}
\label{experiments for k}
\vspace{-20pt}
\end{figure}

\textbf{Influence of LCA based multi-modal fusion.} An uni-modal model (\textit{RGB}) taking only RGB images as input is evaluated as a baseline. We also employ an early fusion baseline following \cite{DBLP:journals/ral/ChuXV18}, \emph{i.e.} \textit{RGD}, which replaces the blue channel with the depth channel. As shown in Table \ref{ablation study on multi-modal fusion}, although jointly using RGB and depth data boosts the performance, the early fusion strategy is sub-optimal because of the distribution gap between two modalities. Furthermore, we evaluate another two middle fusion baselines, which exploit independent feature extractors to process RGB and depth data respectively and sum (\textit{RGB-D-Sum}) or concatenate (\textit{RGB-D-Concat}) the two types of features. The same as in LCA, multi-modal features are fused at the three residual blocks of the ResNet models. The results in Table \ref{ablation study on multi-modal fusion} demonstrate that LCA outperforms the other middle fusion strategies, highlighting its superiority in RGB-D feature fusion for grasp detection. 


To evaluate the influence of the local spatial extent in LCA, we train our network with different combinations of $k_i$ in each block of ResNet, $i=1, 2, 3$. The results are presented in Fig. \ref{experiments for k} and as it shows, the LCA module achieves consistent improvements with different settings of local spatial sizes. However, when the size of the first block is set too large, there is a drop. Based on the ablations, we choose $(k_1, k_2, k_3) = (5, 3, 3)$ to reach the best performance.

\section{Conclusion}

This paper proposes a novel two-stage approach, \emph{i.e.} DGCAN, to RGB-D grasp detection. A complete 6-dimensional rectangle representation is adopted to emphasize the necessity of the grasp depth, which is not considered in the common 5-dimensional definition. The prediction of the grasp depth substantially improves grasp feature learning, thereby leading to more accurate results. Besides, an LCA module is designed, where the depth features are refined according to cross-modal relations and concatenated to the RGB ones for more sufficient fusion. The results of the simulation and physical experiments demonstrate its effectiveness.

\newpage
\bibliographystyle{IEEEtran}
\bibliography{IEEEabrv, IEEEexample}

\end{document}